\documentclass[10pt,twocolumn,letterpaper]{article}
\usepackage{cvpr}
\usepackage{times}
\usepackage{epsfig}
\usepackage{graphicx}
\usepackage{amsmath}
\usepackage{amssymb}
\usepackage{array}
\usepackage{bm}
\usepackage{algorithm}
\usepackage{algorithmic}
\usepackage[labelformat=simple]{subcaption}
\usepackage{multirow}
\usepackage{booktabs}
\newcolumntype{C}[1]{>{\centering\arraybackslash}p{#1}}

\DeclareMathOperator*{\argmin}{arg\,min}

\usepackage[breaklinks=true,bookmarks=false]{hyperref}

\cvprfinalcopy 

\def\cvprPaperID{4122} 

\begin{document}

\title{Single Image Reflection Removal Exploiting Misaligned Training Data and Network Enhancements}

\author{Kaixuan Wei$^1$  \quad Jiaolong Yang$^2$ \quad Ying Fu$^1$ \quad David Wipf$^2$ \quad Hua Huang$^1$
\\ $^1$Beijing Institute of Technology \quad $^2$Microsoft  Research\\
{\tt\small \{kaixuan\_wei,fuying,huahuang\}@bit.edu.cn} \quad
{\tt\small \{jiaoyan,davidwip\}@microsoft.com}
}


\maketitle

\begin{abstract}
Removing undesirable reflections from a single image captured through a glass window is of practical importance to visual computing systems. Although state-of-the-art methods can obtain decent results in certain situations, performance declines significantly when tackling more general real-world cases.  These failures stem from the intrinsic difficulty of single image reflection removal -- the fundamental ill-posedness of the problem, and the insufficiency of densely-labeled training data needed for resolving this ambiguity within learning-based neural network pipelines. In this paper, we address these issues by exploiting targeted network enhancements and the novel use of misaligned data. For the former, we augment a baseline network architecture by embedding context encoding modules that are capable of leveraging high-level contextual clues to reduce indeterminacy within areas containing strong reflections. For the latter, we introduce an alignment-invariant loss function that facilitates exploiting misaligned real-world training data that is much easier to collect.  Experimental results collectively show that our method outperforms the state-of-the-art with aligned data, and that significant improvements are possible when using additional misaligned data. 
\end{abstract}

\section{Introduction}
Reflection is a frequently-encountered source of image corruption that can arise when shooting through a glass surface. Such corruptions can be addressed via the process of single image reflection removal (SIRR), a challenging problem that has attracted considerable attention from the computer vision community~\cite{levin2007user,Li2014Single,wan2016depth,Arvanitopoulos_2017_CVPR,fan2017generic,zhang2018single,eccv18refrmv,Wan_2018_CVPR}. Traditional optimization-based methods often leverage manual intervention or strong prior assumptions to render the problem more tractable~\cite{levin2007user,Li2014Single}. Recently, alternative learning-based approaches rely  on  deep Convectional Neural Networks (CNNs) in lieu of the costly optimization and handcrafted priors~\cite{fan2017generic,zhang2018single,eccv18refrmv,Wan_2018_CVPR}. But promising results notwithstanding, SIRR remains a largely unsolved problem across disparate imaging conditions and varying scene content.

For CNN-based reflection removal, our focus herein, the challenge originates from at least two sources: \textbf{(i)} The extraction of a background image layer devoid of reflection artifacts is fundamentally ill-posed, and \textbf{(ii)} Training data from real-world scenes, are exceedingly scarce because of the difficulty in obtaining ground-truth labels.  

Mathematically speaking, it is typically assumed that a captured image $I$ is formed as a linear combination of a background or transmitted layer $T$ and a reflection layer $R$, \ie, $I=T+R$.  Obviously, when given access only to $I$, there exists an infinite number of feasible decompositions.  Further compounding the problem is the fact that both $T$ and $R$ involve content from real scenes that may have overlapping appearance distributions. This can make them difficult to distinguish even for human observers in some cases, and simple priors that might mitigate this ambiguity are not available except under specialized conditions.

On the other hand, although CNNs can perform a wide variety visual tasks, at times exceeding human capabilities, they generally require a large volume of labeled training data. Unfortunately, 
real reflection images accompanied with densely-labeled, ground-truth transmitted layer intensities are scarce. Consequently, previous learning-based approaches have resorted to training with synthesized images \cite{fan2017generic,Wan_2018_CVPR,zhang2018single} and/or small real-world data captured from specialized devices~\cite{zhang2018single}. However, existing image synthesis procedures are heuristic and the domain gap may jeopardize accuracy on real images. On the other hand, collecting sufficient additional real data with precise ground-truth labels is tremendously labor-intensive.

This paper is devoted to addressing both of the aforementioned challenges.  First, to better tackle the intrinsic ill-posedness and diminish ambiguity, we propose to leverage a network architecture that is sensitive to contextual information, which has proven useful for other vision tasks such as semantic segmentation~\cite{He2015Spatial,zhao2017pyramid,Zhang_2018_CVPR,Hu_2018_CVPR}. Note that at a high level, our objective is to efficiently convert prior information mined from labeled training data into network structures capable of resolving this ambiguity.  Within a traditional CNN model, especially in the early layers where the effective receptive field is small, the extracted features across all channels are inherently local.  However, broader non-local context is necessary to differentiate those features that are descriptive of the desired transmitted image, and those that can be discarded as reflection-based.  For example, in image neighborhoods containing a particularly strong reflection component, accurate separation by any possible method (even one trained with arbitrarily rich training data) will likely require contextual information from regions without reflection.  To address this issue, we utilize two complementary forms of context, namely, channel-wise context and multi-scale spatial context. Regarding the former, we apply a channel attention mechanism to the feature maps from convolutional layers such that different features are weighed differently according to global statistics of the activations. For the latter, we aggregate information across a pyramid of feature map scales within each channel to reach a global contextual consistency in the spatial domain. Our experiments demonstrate that significant improvement can be obtained by these enhancements, leading to state-of-the-art performance on two real-image datasets. 

\begin{figure}[t]
	\centering
	\setlength\tabcolsep{1pt}
	\begin{tabular}{cc}
		\includegraphics[width=0.41\columnwidth]{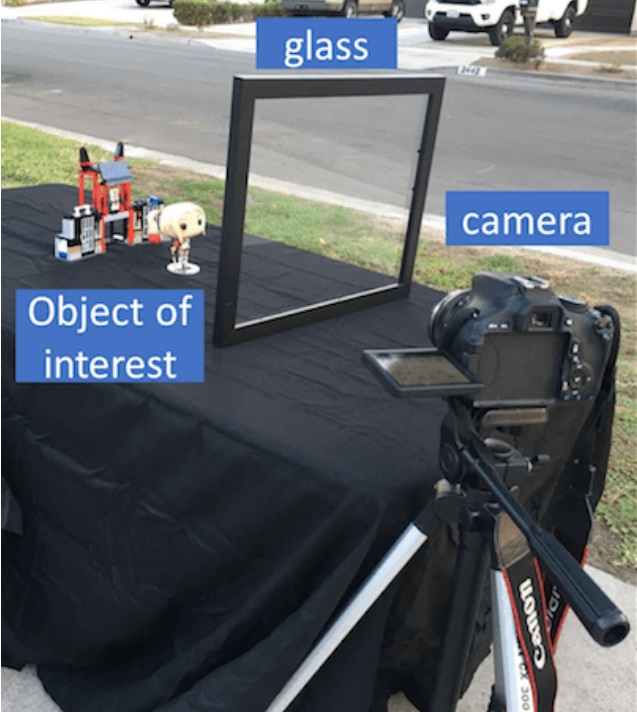} & \includegraphics[width=0.436\columnwidth]{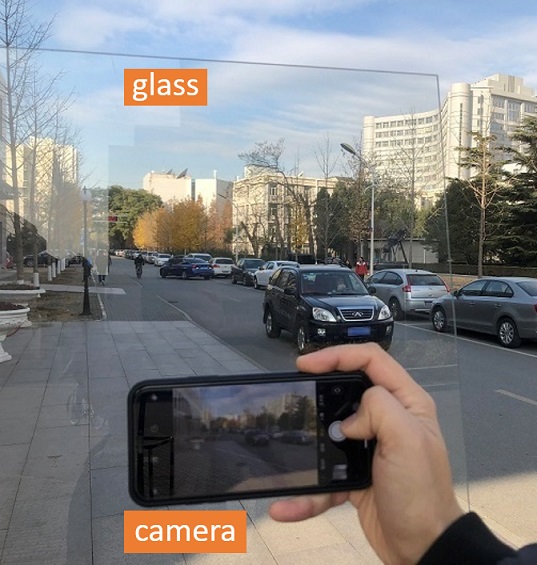}\\
		\cite{Zhang_2018_CVPR} & Ours
	\end{tabular}
	\vspace{-7pt}
	\caption{Comparison of the reflection image data collection methods in \cite{Zhang_2018_CVPR} and this paper.}\label{fig:datacollection}
	\vspace{-8pt}
\end{figure}

Secondly, orthogonal to architectural considerations, we seek to expand the sources of viable training data by facilitating the use of misaligned training pairs, which are considerably easier to collect.  Misalignment between an input image $I$ and a ground-truth reflection-free version $T$ can be caused by camera and/or object movements during the acquisition process.  In the previous works~\cite{Wan_2017_ICCV,Zhang_2018_CVPR}, data pairs $(I,T)$ were obtained by taking an initial photo through a glass plane, followed by capturing a second one after the glass has been removed.  This process requires that the camera, scene, and even lighting conditions remain static. Adhering to these requirements across a broad acquisition campaign can significantly reduce both the quantity and diversity of the collected data. Additionally, post-processing may also be necessary to accurately align $I$ and $T$ to compensate for spatial shifts caused by the refractive effect~\cite{Wan_2017_ICCV}. In contrast, capturing unaligned data is considerably less burdensome, as shown in Fig.~\ref{fig:datacollection}.  For example, there is no need for a tripod, table, or other special hardware; the camera can be hand-held and the pose can be freely adjusted; dynamic scenes in the presence of vehicles, humans, \etc can be incorporated; and finally no post-processing of any type is needed.


To handle such misaligned training data, we require a loss function that is, to the extent possible, invariant to the alignment, \ie, the measured image content discrepancy between the network prediction and its unaligned reference should be similar to what would have been observed if the reference was actually aligned.  In the context of image style transfer~\cite{johnson2016perceptual} and others, certain perceptual loss functions have been shown to be relatively invariant to various transformations. Our study shows that the using only the highest-level feature from a deep network (VGG-19 in our case) leads to satisfactory results for our reflection removal task.
In both simulation tests and experiments using a newly collected dataset, we demonstrate for the first time that training/fine-tuning a CNN with unaligned data improves the reflection removal results by a large margin.

\section{Related Work}

This paper is concerned with reflection removal from a single image. Previous methods utilizing multiple input images of, \eg, flash/non-flash pairs~\cite{agrawal2005removing}, different polarization~\cite{kong2014physically}, multi-view or video sequences~\cite{farid1999separating,szeliski2000layer,sarel2004separating,gai2012blind,li2013exploiting,sinha2012image,guo2014robust,xue2015computational,yang2016robust} will not be considered here. 


\vspace{5pt}
\noindent\textbf{Traditional methods.}
Reflection removal from a single image is a massively ill-posed problem. Additional priors are needed to solve the otherwise prohibitively-difficult problem in traditional optimization-based method~\cite{levin2007user,Li2014Single,wan2016depth,Arvanitopoulos_2017_CVPR,Wan2018Region}. In \cite{levin2007user}, user annotations are used to guide layer separation jointly with a gradient sparsity prior~\cite{levin2003learning}. \cite{Li2014Single} introduces a relative smoothness prior where the reflections are assumed to be blurry thus their large gradients are penalized. \cite{wan2016depth} explores a variant of the smoothness prior where a multi-scale Depth-of-Field (DoF) confidence map is utilized to perform edge classification. \cite{Shih_2015_CVPR} exploits the ghost cues for layer separation. \cite{Arvanitopoulos_2017_CVPR} proposes a simple optimization formulation with an $l_0$ gradient penalty on the transmitted layer inspired by image smoothing algorithms~\cite{L0smooth}.
Despite decent results can be obtained by these methods where their assumptions hold, the vastly-different imaging conditions and complex scene content in the real world render their generalization problematic.

\vspace{5pt}
\noindent\textbf{Deep learning based methods.~}
Recently, there is an emerging interest in applying deep convolutional neural networks for single image reflection removal such that the handcrafted priors can be replaced by data-driven learning~\cite{fan2017generic,Wan_2018_CVPR,zhang2018single,eccv18refrmv}. 
The first CNN-based method is due to \cite{fan2017generic}, where a network structure is proposed to first predict the background layer in the edge domain followed by reconstructing it the color domain. Later, \cite{Wan_2018_CVPR} proposes to predict the edge and image intensity concurrently by two cooperative sub-networks. The recent work of \cite{eccv18refrmv} presents a cascade network structure which predicts the background layer and reflection layer in an interleaved fashion. The earlier CNN-based methods typical use the raw image intensity discrepancy such as mean squared error (MSE) to train the networks. Several recent works~\cite{zhang2018single,jin2018learning,chi2018single} adopt the perceptual loss~\cite{johnson2016perceptual} which uses the multi-stage features of a deep network pre-trained on ImageNet~\cite{russakovsky2015imagenet}.
\cite{zhang2018single}. Adversarial loss is investigated in \cite{zhang2018single,lee2018generative} to improve the realism of the predicted background layers.

\begin{figure*}
\vspace{-5pt}
	\centering
     \includegraphics[width=.99\linewidth]{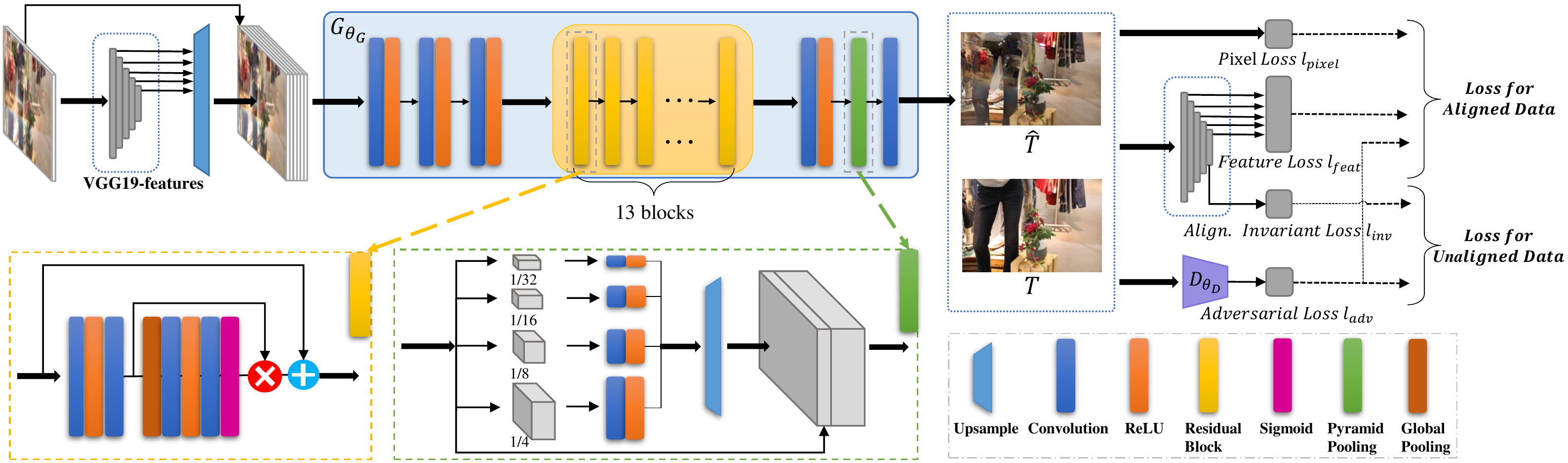}
     \vspace{-2pt}
	\caption{Overview of our approach for single image reflection removal.}\label{fig:method_overview}
\vspace{-1pt}
\end{figure*}

\section{Approach} \label{sec:approach}
Given an input image $I$ contaminated with reflections, our goal is to estimate a reflection-free trasmitted image $\hat{T}$.
To achieve this, we train a feed-forward CNN $G_{\theta_G}$ parameterized by $\theta_G$ to minimize a reflection removal loss function $l$. Given training image pairs $\{(I_n, T_n)\}$, $n = 1, \cdots, N$, this involves solving:
\begin{align}
\textstyle{\hat{\theta}_G =  \argmin_{\theta_G} \frac{1}{N} \sum_{n=1}^{N} l (G_{\theta_G} (I_n), T_n)}.
\end{align}
We will first introduce the details of network architecture $G_{\theta_G}$ followed by the loss function $l$ applied to both aligned data (the common case) and newly proposed unaligned data extensions.  The overall system is illustrated in Fig.~\ref{fig:method_overview}.

\subsection{Basic Image Reconstruction Network} \label{sec:eicnn}

Our starting point can be viewed as the basic image reconstruction neural network component from \cite{fan2017generic} but modified in three aspects: (1) We simplify the basic residual block \cite{He_2016_CVPR} by removing the batch normalization (BN) layer \cite{ioffe2015batch}; (2) we increase the  capacity by widening the network from 64 to 256 feature maps; and (3)  for each input image $I$,  we extract hypercolumn features \cite{Hariharan_2015_CVPR}  from a pretrained VGG-19 network \cite{DBLP:journals/corr/SimonyanZ14a}, and concatenate these features with $I$ as an augmented network input.  As explained in \cite{zhang2018single}, such an augmentation strategy can help enable the network to learn semantic clues from the input image.

Note that removing the BN layer from our network  turns out to be critical for optimizing performance in the present context.  As shown in \cite{wu2018group}, if batch sizes become too small, prediction errors can increase precipitously and stability issues can arise.  Moreover, for a dense prediction task such as SIRR, large batch sizes can become prohibitively expensive in terms of memory requirements.  In our case, we found that within the tenable batch sizes available for reflection removal, BN led to considerably worse performance, including color attenuation/shifting issues as sometimes observed in image-to-image translation tasks \cite{fan2017generic,Isola_2017_CVPR,Zhu_2017_ICCV}.  BN layers have similarly been removed from other dense prediction tasks such as image super-resolution \cite{Lim_2017_CVPR_Workshops} or deblurring \cite{Nah_2017_CVPR}.



At this point, we have constructed a useful base architecture upon which other more targeted alterations will be applied shortly.  This baseline, which we will henceforth refer to as \emph{BaseNet},  performs quite well when trained and tested on synthetic data.  However, when deployed on real-world reflection images we found that its performance degraded by an appreciable amount, especially on the 20 real images from \cite{zhang2018single}.  Therefore, to better mitigate the transition from the make-believe world of synthetic images to real-life photographs, we describe two modifications for introducing broader contextual information into otherwise local convolutional filters.


\subsection{Context Encoding Modules}

As mentioned previously, we consider both context between channels and multi-scale context within channels.

\vspace{5pt}
{\noindent \bf Channel-wise context.} 
The underlying design principle here is to introduce global contextual information across channels, and a richer overall structure within residual blocks, without dramatically increasing the parameter count.  One way to accomplish this is by incorporating a channel attention module originally developed in \cite{Hu_2018_CVPR} to recalibrate feature maps using global summary statistics.  


Let  $U =  [u_1, \ldots, u_c, \ldots, u_C ]$ denote original, uncalibrated activations produced by a network block, with $C$ feature maps of size of $H \times W$.  These activations generally only reflect local information residing within the corresponding receptive fields of each filter.  We then form scalar, channel-specific descriptors $z_c = f_{gp}(u_c)$ by applying a global average pooling operator $f_{gp}$ to each feature map $u_c \in \mathbb{R}^{H\times W}$. The vector $z = [z_1, \ldots, z_C] \in \mathbb{R}^C$ represents a simple statistical summary of global, per-channel activations and, when passed through a small network structure, can be used to adaptively predict the relative importance of each channel \cite{Hu_2018_CVPR}.  

More specifically, the channel attention module first computes
$s = \sigma (W_U \delta (W_D z))$
where $W_D$ is a trainable weight matrix that downsamples $z$ to dimension $R < C$, $\delta$ is a ReLU non-linearity, $W_U$ represents a trainable upsampling weight matrix, and $\sigma$ is a sigmoidal activation.  Elements of the resulting output vector $s \in \mathbb{R}^C$ serve as channel-specific gates for calibrating feature maps via
$\hat{u}_c = s_c \cdot u_c$.

Consequently, although each individual convolutional filter has a local receptive field, the determination of which channels are actually important in predicting the transmission layer and suppressing reflections is based on the processing of a global statistic (meaning the channel descriptors computed as activations pass through the network during inference).  Additionally, the parameter overhead introduced by this process is exceedingly modest given that $W_D$ and $W_U$ are just small additional weight matrices associated with each block.



\vspace{5pt}
{\noindent \bf Multi-scale spatial context.}
Although we have found that encoding the contextual information across channels already leads to significant empirical gains on real-world images, utilizing complementary multi-scale spatial information within each channel provides further benefit.  To accomplish this, we apply a pyramid pooling module \cite{He2015Spatial}, which has proven to be an effective global-scene-level representation in semantic segmentation \cite{zhao2017pyramid}.  As shown in Fig.~\ref{fig:method_overview}, we construct such a module using pooling operations at sizes 4, 8, 16, and 32 situated in the tail of our network before the final construction of $\hat{T}$.  Pooling in this way fuses features under four different pyramid scales. After harvesting the resulting sub-region representations, we perform a non-linear transformation (\ie a Conv-ReLU pair) to reduce the channel dimension. The refined features are then upsampled via bilinear interpolation. Finally, the different levels of features are concatenated together as a final representation reflecting multi-scale spatial context within each channel; the increased parameter overhead is negligible.


\subsection{Training Loss for Aligned Data}


In this section, we present our loss function for aligned training pairs $(I,T)$, which consists of three terms similar to previous methods~\cite{zhang2018single,eccv18refrmv}.

\vspace{5pt}
\noindent\textbf{Pixel loss.~}
Following \cite{fan2017generic}, we penalize the pixel-wise intensity difference of $T$ and $\hat{T}$ via $l_{pixel}=\alpha \|\hat{T}-T\|^2_2 + \beta (\|\nabla_x \hat{T}-\nabla_x T\|_1 + \|\nabla_y \hat{T}-\nabla_y T\|_1)$
where $\nabla_x$ and $\nabla_y$ are the gradient operator along x- and y-direction, respectively. We set $\alpha=0.2$ and $\beta=0.4$ in all our experiments.

\vspace{5pt}
\noindent\textbf{Feature loss.~} 
We define the feature loss based on the activations of the 19-layer VGG network~\cite{simonyan2014very} pretrained on ImageNet~\cite{russakovsky2015imagenet}. Let $\phi_l$ be the feature from the $l$-th layer of VGG-19, we define the feature loss as $l_{feat} = \sum_l \lambda_l \| \phi_l (T) - \phi_l (\hat{T}) \|_1$
where $\{ \lambda_l \}$ are the balancing weights. Similar to \cite{zhang2018single}, we use the layers `conv2\_2', `conv3\_2', `conv4\_2', and  `conv5\_2' of VGG-19 net.


\vspace{6pt}
\noindent\textbf{Adversarial loss.~}
We further add an adversarial loss to improve the realism of the produced background images. We define an opponent discriminator network $D_{\theta_D}$ and minimize the relativistic adversarial loss \cite{jolicoeurmartineau2018the} defined as $l_{adv} = l^{G}_{adv} = - \log (D_{\theta_D} (T, \hat{T}))  
- \log (1 - D_{\theta_D} (\hat{T}, T))$ for $G_{\theta_G}$ and $l^{D}_{adv} = - \log (1 - D_{\theta_D} (T,  \hat{T})) - \log (D_{\theta_D} (\hat{T}, T))$ for $D_{\theta_D}$ where $D_{\theta_D} (T, \hat{T}) =\sigma (C(T) - C(\hat{T}))$ with  $\sigma(\cdot)$ being the sigmoid function and $C(\cdot) $ the non-transformed discriminator function (refer to \cite{jolicoeurmartineau2018the} for details). 



To summarize, our loss for aligned data is defined as:
\begin{align}
l_{aligned} = \omega_1 l_{pixel} + \omega_2 l_{feat}  + \omega_3 l_{adv}
\label{eq:alignedLoss}
\end{align} 
where we empirically set the weights as $\omega_1 = 1, \omega_2 = 0.1$, and $\omega_3 = 0.01$ respectively throughout our experiments.
\subsection{Training Loss for Unaligned Data}

To use misaligned data pairs $(I,T)$ for training, we need a loss function that is invariant to the alignment, such that the true similarity between $T$ and the prediction $\hat{T}$ can be reasonably measured. In this regard, we note that human observers can easily assess the similarity of two images even if they are not aligned. Consequently, designing a loss measuring image similarity on the perceptual-level may serve our goal. This motivates us to directly use a deep feature loss for unaligned data. 

\begin{figure}
\centering
\setlength\tabcolsep{1pt}
\begin{tabular}{ccc}
    \includegraphics[width=0.3\linewidth]{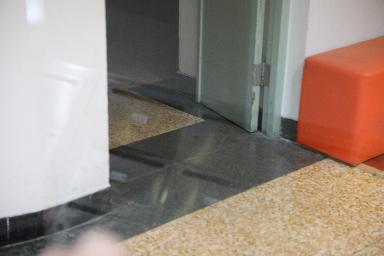}
	& \includegraphics[width=0.3\linewidth]{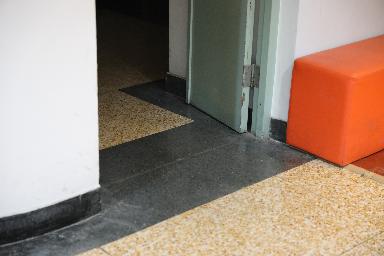} 
	& \includegraphics[width=0.3\linewidth]{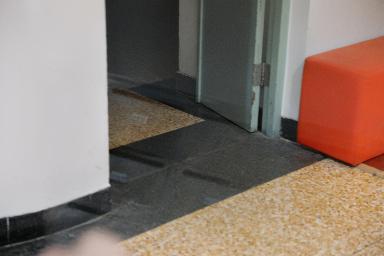} \vspace{-3pt}\\
	(a) Input &  (b) Unaligned Ref. & (c) Pretrained \vspace{2pt}\\
	\includegraphics[width=0.3\linewidth]{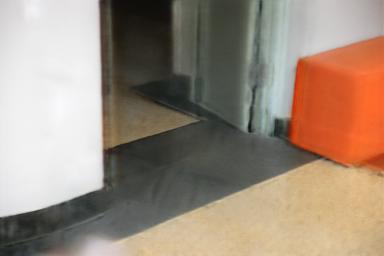}
	& \includegraphics[width=0.3\linewidth]{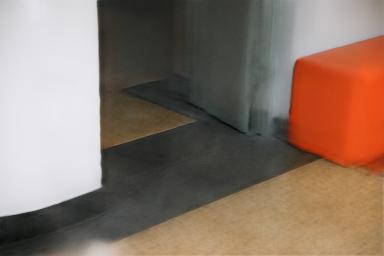} 
	& \includegraphics[width=0.3\linewidth]{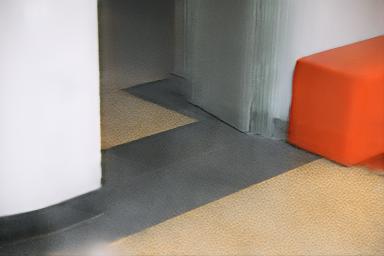} \vspace{-3pt}\\
	(d) $l_{pixel}$ & (e) conv2\_2 & (f) conv3\_2 \vspace{2pt}\\
	\includegraphics[width=0.3\linewidth]{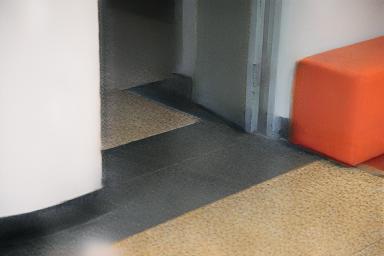}
	& \includegraphics[width=0.3\linewidth]{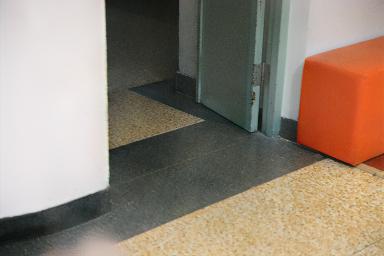} 
	& \includegraphics[width=0.3\linewidth]{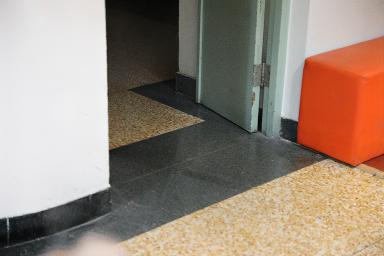} \vspace{-3pt}\\
	(g) conv4\_2 &  (h) \textbf{conv5\_2} & (i) Loss of \cite{Mechrez_2018_ECCV}
\end{tabular}
\caption{The effect of using different loss to handle misaligned real data. (a) and (b) are the unaligned image pair $(I,T)$. (c) shows the reflection removal result of our network trained on synthetic data and a small number of aligned real data (see Section~\ref{sec:experiments} for details). Reflection can still be observed in the predicted background image. (d) is the result finetuned on $(I,T)$ with pixelwise intensity loss. (e)-(h) are the results finetuned with features at different layers of VGG-19. Only the highest-level feature from `conv5\_2' yields satisfactory result. (i) shows the results finetuned with the loss of \cite{Mechrez_2018_ECCV}. (\textbf{Best viewed on screen with zoom})
}
\label{fig:real-world-example}
\end{figure}

Intuitively, the deeper the feature, the more likely it is to be insensitive to misalignment. To experimentally verify this and find a suitable feature layer for our purposes, we conducted tests using a pre-trained VGG-19 network as follows. Given an unaligned image pair $(I,T)$, we use gradient descent to finetune the weights of our network $G_{\theta_G}$ to minimize the feature difference of $T$ and $\hat{T}$, with features extracted at different layers of VGG-19. Figure~\ref{fig:real-world-example} shows that using low-level or middle-level features from `conv2\_2' to `conv4\_2' leads to blurry results (similar to directly using a pixel-wise loss), although the reflection is more thoroughly removed. In contrast, using the highest-level feature from `conv5\_2' gives rise to a striking result: the predicted background image is sharp and almost reflection-free.

Recently, \cite{Mechrez_2018_ECCV} introduced a ``contextual loss" which is also designed for training deep networks with unaligned data for image-to-image translation tasks like image style transfer. In Fig~\ref{fig:real-world-example}, we also present the finetuned result using this loss for our reflection removal task. Upon visual inspection, the results are similar to our highest-level VGG feature loss (quantitative comparison can be found in the experiment section). However, our adopted loss (formally defined below) is much simpler and more computationally efficient than the loss from \cite{Mechrez_2018_ECCV}.

\vspace{6pt}
\noindent\textbf{Alignment-invariant loss.~}
Based on the above study, we now formally define our invariant loss component designed for unaligned data as $l_{inv} = \| \phi_h (T) - \phi_h (\hat{T}) \|_1$, where $\phi_h$ denotes the `conv5\_2' feature of the pretrained VGG-19 network.  For unaligned data, we also apply an adversarial loss which is not affected by misalignment. Therefore, our overall loss for unaligned data can be written as
\begin{align}
l_{unaligned} = \omega_4 l_{inv}  + \omega_5 l_{adv}
\label{eq:zz}
\end{align}
where we set the weights as $\omega_4 = 0.1$ and $\omega_5 = 0.01$.

\section{Experiments}\label{sec:experiments}

 
\subsection{Implementation Details} \label{sec:implement}
\noindent{\bf Training data.} We adopt a fusion of synthetic and real data as
our train dataset. The images from  \cite{fan2017generic} are used as
sythetic data, \ie 7,643 cropped images with size $224 \times 224$ from PASCAL
VOC dataset \cite{everingham2010pascal}. 90 real-world training images from
\cite{zhang2018single} are adopted as real data. For image synthesis, we use the
same data generation model as \cite{fan2017generic} to create our synthetic
data. In the following, we always use the same dataset for training, unless
specifically stated. 

\vspace{5pt}
\noindent {\bf Training details.} Our implementation\footnote{Code is released on https://github.com/Vandermode/ERRNet} is based on 
PyTorch. 
We train the model with 60
epoch using the Adam optimizer \cite{kingma2014adam}. The
base learning rate is set to $10^{-4}$ and halved at epoch 30, then reduced to
$10^{-5}$ at epoch 50. The weights are initialized as in
\cite{Lim_2017_CVPR_Workshops}.
 
\subsection{Ablation Study} \label{sec:com-analysis}


In this section, we conduct an ablation study for our method on 100 synthetic testing images from
 \cite{fan2017generic} and 20 real testing images from 
\cite{zhang2018single} (denoted by `Real20'). 


\vspace{5pt}
{\noindent \bf Component analysis.~} To verify the importance of our network
design, we compare four model architectures as described in Section
\ref{sec:approach}, including (1) Our basic image reconstruction network
\emph{BaseNet}; (2) \emph{BaseNet} with channel-wise context module (\emph{BaseNet} + CWC); (3) \emph{BaseNet} with multi-scale spatial context module (\emph{BaseNet} + MSC); and (4) Our enhanced reflection removal
network, denoted ERRNet,
\ie, \emph{BaseNet} + CWC + MSC. 
The result from the CEILNet \cite{fan2017generic} fine-tuned on our training
data (denoted by CEILNet-F) is also provided as an additional reference.

\begin{figure}[t]
	\centering
	\small
	\setlength\tabcolsep{1pt}
	\begin{tabular}{ccc}
			Input & BaseNet   & ERRNet \\
        	 \includegraphics[width=0.3\linewidth]{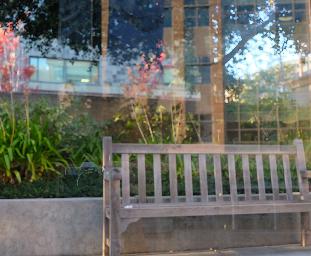}
		& \includegraphics[width=0.3\linewidth]{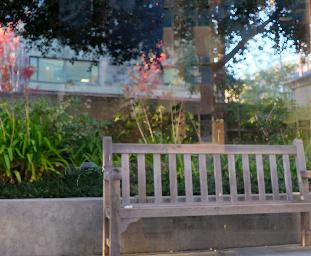} 
		& \includegraphics[width=0.3\linewidth]{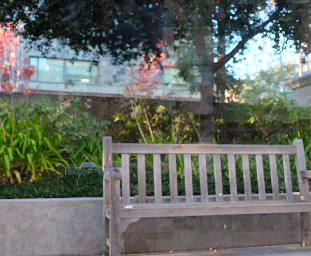} \\
		     \includegraphics[width=0.3\linewidth]{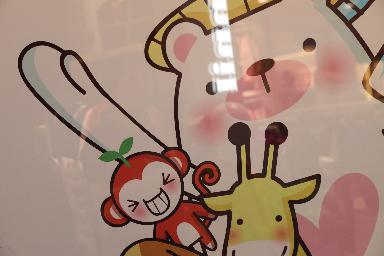}
		& \includegraphics[width=0.3\linewidth]{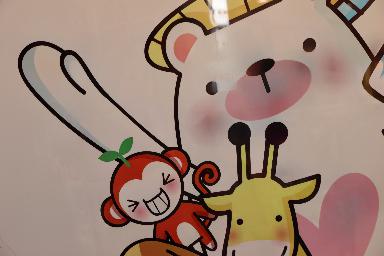} 
		& \includegraphics[width=0.3\linewidth]{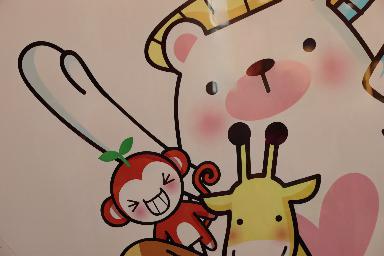} 
	\end{tabular}
	\caption{Comparison of the results with (ERRNet) and without (BaseNet) the context encoding modules.}
	\label{fig:intuitive-example}
\end{figure}

As shown in Table \ref{tb:componet}, our \emph{BaseNet} has already achieved a
much better result than CEILNet-F.
The performance of our \emph{BaseNet} could be obviously boosted by using
channel-wise context and multi-scale spatial context modules, especially by using them
together, \ie ERRNet. 
Figure \ref{fig:intuitive-example} visually shows the results from BaseNet and
our ERRNet. It can be observed that \emph{BaseNet} struggles to discriminate the
reflection region and yields some obvious residuals, while the ERRNet removes the reflection and produces much cleaner transmitted images. 
These results suggest the effectiveness of our network design, especially the components tailored to encode the contextual clues.

\begin{table}[!t]
\centering
\caption{Comparison of different settings. Our full model (\ie ERRNet) leads to best performance among all comparisons.}
\footnotesize
\begin{tabular}{lcccc} 
		\toprule
		 & \multicolumn{2}{c}{Synthetic} & \multicolumn{2}{c}{Real20} \\
		Model & PSNR & SSIM & PSNR & SSIM \\ 
		\midrule
		CEILNet-F \cite{fan2017generic}  & 24.70 & 0.884 & 20.32 & 0.739 \\ \hline
		\midrule
		BaseNet only & 25.71 &  0.926 &21.51 & 0.780 \\ \hline	
		BaseNet + CSC & 27.64 & 0.940  & 22.61 & 0.796 \\ \hline	
		BaseNet + MSC & 26.03 & 0.928 & 21.75 & 0.783 \\ \hline
		ERRNet & \textbf{27.88} & \textbf{0.941} & \textbf{22.89}
 & \textbf{0.803} \\
		\bottomrule
\end{tabular}
\label{tb:componet}
\end{table}

\vspace{5pt}
\noindent {\bf Efficacy of the training loss for unaligned data.~}
In this experiment, we first train our ERRNet with only `synthetic data', `synthetic + 50
aligned real data', and `synthetic + 90 aligned real data'. The loss
function in Eq.~\eqref{eq:alignedLoss} is used for aligned data. We can
see that the testing results become better with the increasing real data in
Table \ref{tb:unalignedLoss}. 

Then, we synthesize misalignment through
performing random translations within $[-10, 10]$ pixels on real data\footnote{Our alignment-invariant loss $l_{inv}$ can handle shifts of up to 20 pixels.  See \textit{suppl. material} for more details.}, and train
ERRNet with `synthetic + 50 aligned real data + 40 unaligned data'. Pixel-wise loss
$l_{pixel}$ and alignment-invariant
loss $l_{inv}$ are used for 40 unaligned
images. Table \ref{tb:unalignedLoss} shows employing 40 unaligned data with
$l_{pixel}$ loss degrades the performance, even worse than that from 50 aligned
images without additional unaligned data.

In addition, we also investigate the contextual loss $l_{cx}$ of
\cite{Mechrez_2018_ECCV}.  
Results from both contextual loss $l_{cx}$ and our 
alignment-invariant loss
$l_{inv}$ (or combination of them $l_{inv} + l_{cx}$) surpass analogous results obtained with only aligned images by appreciable margins, indicating that these losses provide useful supervision
to networks granted unaligned data. Note although $l_{inv}$ and $l_{cx}$ perform 
equally well, our $l_{inv}$ is much simpler and computationally efficient than
$l_{cx}$, suggesting $l_{inv}$ is lightweight alternative to $l_{cx}$ in terms of
our reflection removal task.

\begin{table}[t]
\centering
\caption{Simulation experiment to verify the efficacy our alignment-invariant loss}
\footnotesize
\begin{tabular}{lcc} 
		\toprule
		Training Scheme & PSNR & SSIM  \\ 
		\midrule
		Synthetic only & 19.79 & 0.741 \\ \hline
		+ 50 aligned & 22.00 &  0.785 \\ \hline		
		+ 90 aligned & 22.89 & 0.803 \\ \hline		
		\midrule
		\multicolumn{3}{c}{+ 50 aligned, + 40 unaligned trained with:} \vspace{3pt}\\
		$l_{pixel}$ & 21.85 &  0.766 \\ \hline
		$l_{inv} $ & 22.38 & \textbf{0.797}   \\ \hline		
		$l_{cx}$ & \textbf{22.47} &  0.796 \\ \hline		
		$l_{inv} $ + $l_{cx}$ & 22.43 & 0.796 \\ \hline
		\bottomrule
\end{tabular}
\label{tb:unalignedLoss}
\end{table}

\begin{figure*}[!t]
	\centering
	\setlength\tabcolsep{1pt}

	\begin{tabular}{ccccccc}
		Input  & LB14 \cite{Li2014Single} & CEILNet-F \cite{fan2017generic} & Zhang \etal \cite{zhang2018single} & BDN-F \cite{eccv18refrmv} & ERRNet & Reference \\
        	 \includegraphics[width=0.142\linewidth]{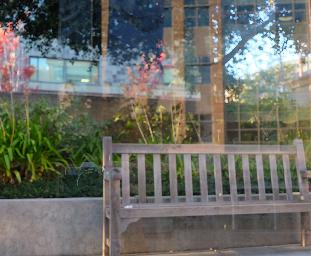}
		&  \includegraphics[width=0.142\linewidth]{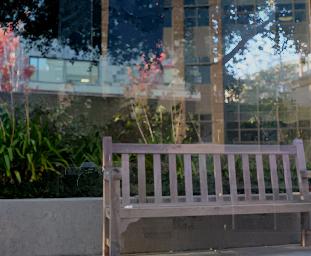}
		&  \includegraphics[width=0.142\linewidth]{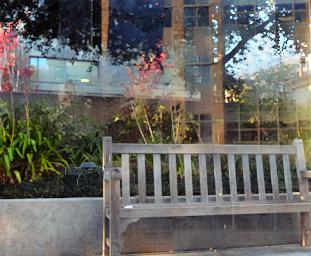}
		&  \includegraphics[width=0.142\linewidth]{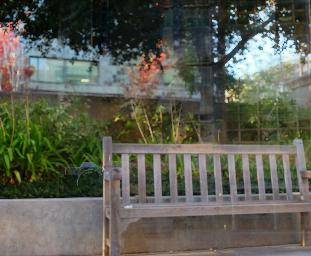}
		&  \includegraphics[width=0.142\linewidth]{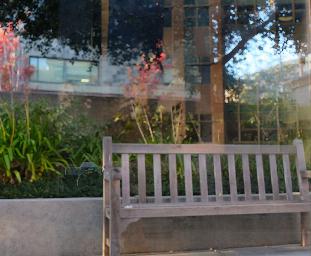}
		&  \includegraphics[width=0.142\linewidth]{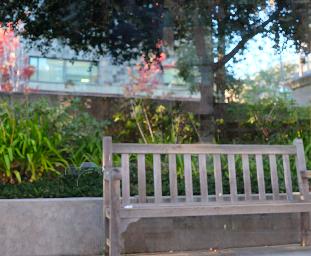}	 
		&  \includegraphics[width=0.142\linewidth]{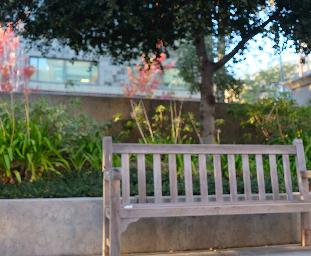}        	 		\\		
        	 \includegraphics[width=0.142\linewidth]{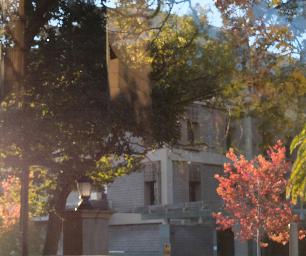}
		&  \includegraphics[width=0.142\linewidth]{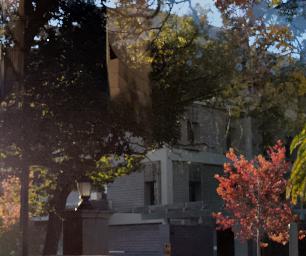}
		&  \includegraphics[width=0.142\linewidth]{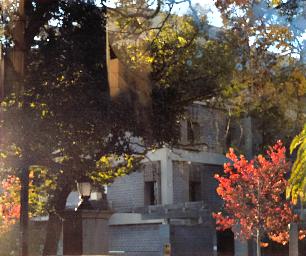}
		&  \includegraphics[width=0.142\linewidth]{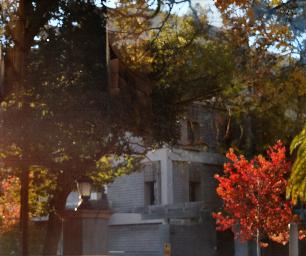}
		&  \includegraphics[width=0.142\linewidth]{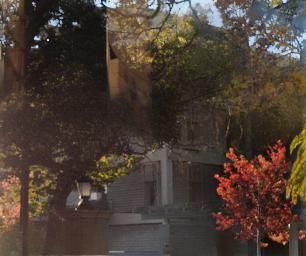}
		&  \includegraphics[width=0.142\linewidth]{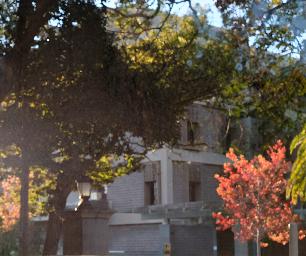}	 
		&  \includegraphics[width=0.142\linewidth]{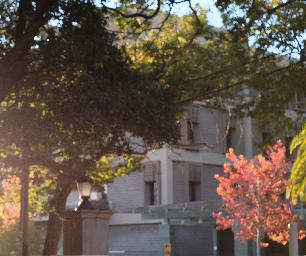}        	 		\\	
        	 \includegraphics[width=0.142\linewidth]{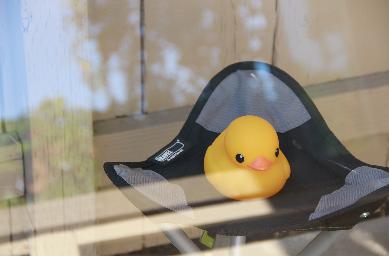}
		&  \includegraphics[width=0.142\linewidth]{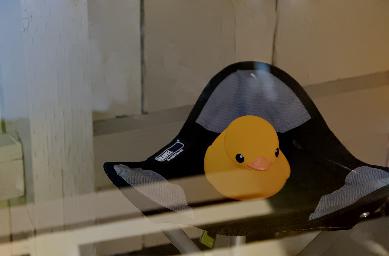}
		&  \includegraphics[width=0.142\linewidth]{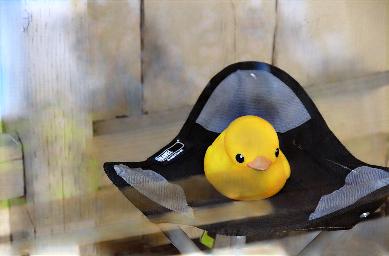}
		&  \includegraphics[width=0.142\linewidth]{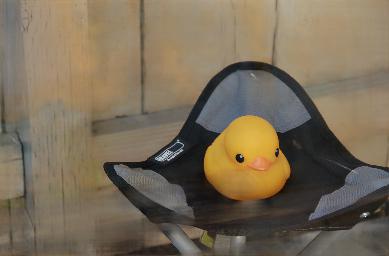}
		&  \includegraphics[width=0.142\linewidth]{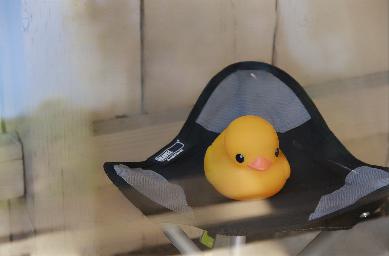}
		&  \includegraphics[width=0.142\linewidth]{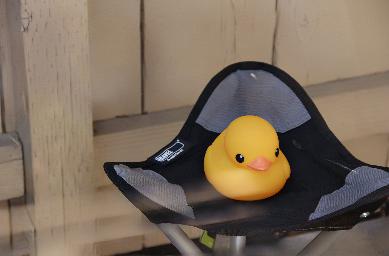}	
		&  \includegraphics[width=0.142\linewidth]{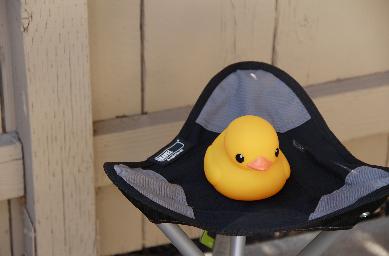}        	 		 \\			
        	 \includegraphics[width=0.142\linewidth]{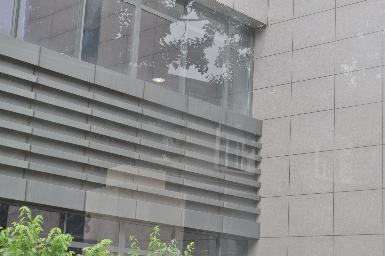}
		&  \includegraphics[width=0.142\linewidth]{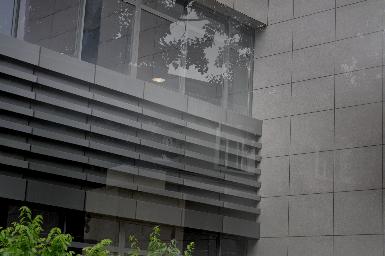}
		&  \includegraphics[width=0.142\linewidth]{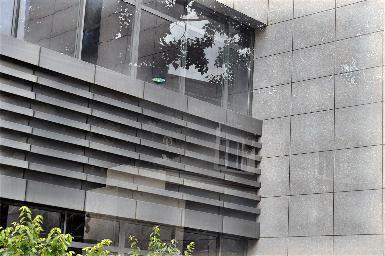}
		&  \includegraphics[width=0.142\linewidth]{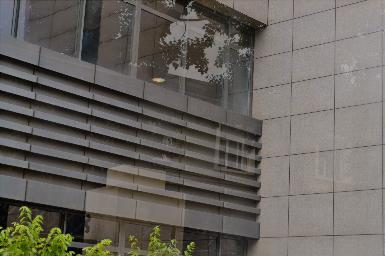}
		&  \includegraphics[width=0.142\linewidth]{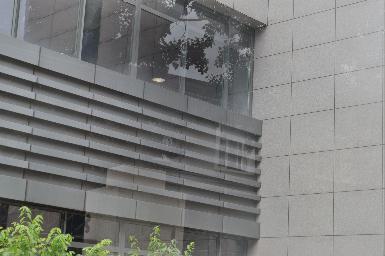}
		&  \includegraphics[width=0.142\linewidth]{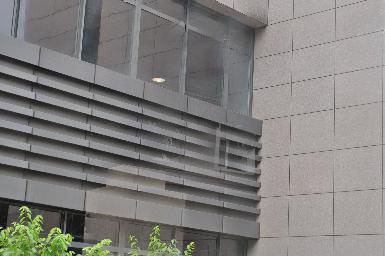}	 
		&  \includegraphics[width=0.142\linewidth]{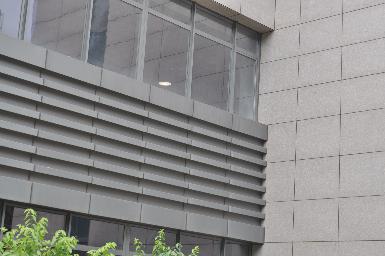}        	 		\\
      	 \includegraphics[width=0.142\linewidth]{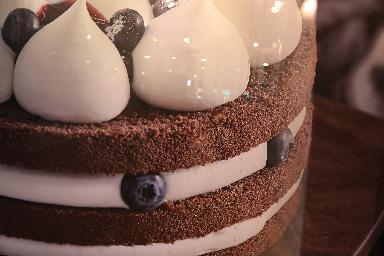} 
		&  \includegraphics[width=0.142\linewidth]{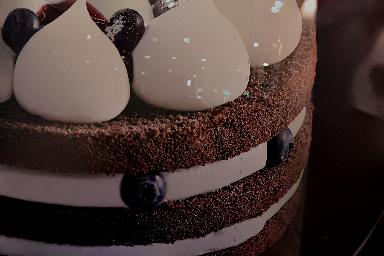}
		&  \includegraphics[width=0.142\linewidth]{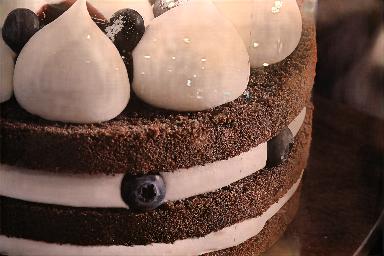}
		&  \includegraphics[width=0.142\linewidth]{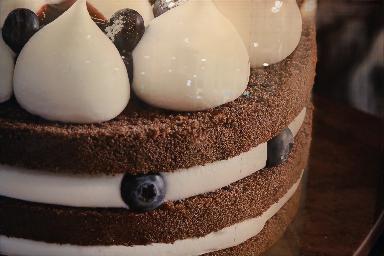}
		&  \includegraphics[width=0.142\linewidth]{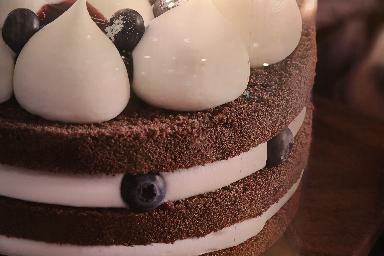}
		&  \includegraphics[width=0.142\linewidth]{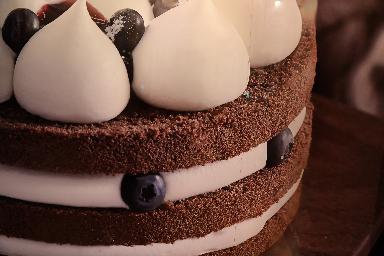}	 
		&  \includegraphics[width=0.142\linewidth]{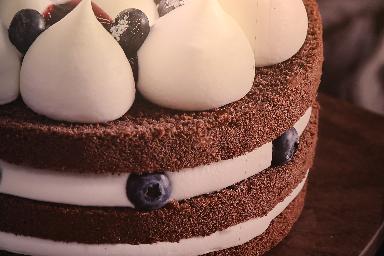}        			\\
      	 \includegraphics[width=0.142\linewidth]{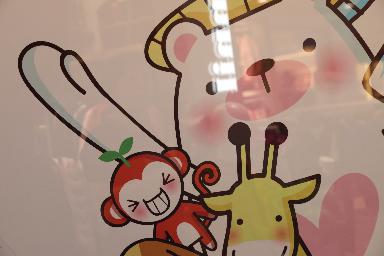}	 
		&  \includegraphics[width=0.142\linewidth]{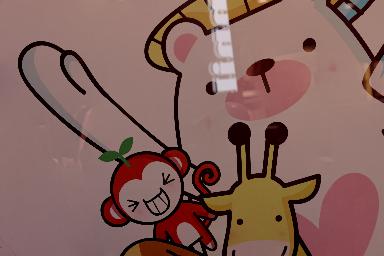}
		&  \includegraphics[width=0.142\linewidth]{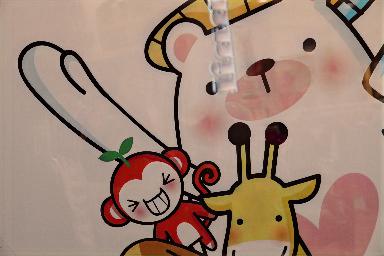}
		&  \includegraphics[width=0.142\linewidth]{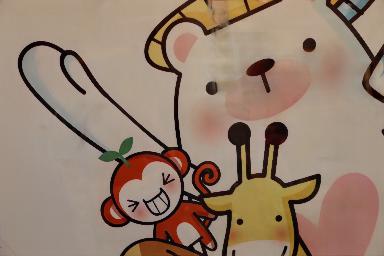}
		&  \includegraphics[width=0.142\linewidth]{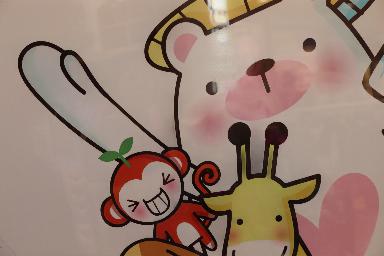}
		&  \includegraphics[width=0.142\linewidth]{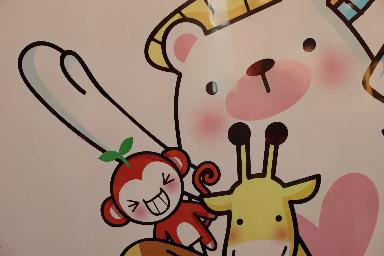}	
		&  \includegraphics[width=0.142\linewidth]{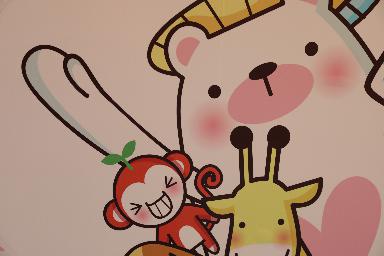}        		 \\
	\end{tabular} 
	\caption{Visual comparison on real-world images. The images are
      obtained from `Real20' (Rows 1-3) and our collected unaligned dataset (Rows 4- 6). More results can be found in the \emph{suppl. material}.}
	\label{fig:method-comparision}
\end{figure*}

\begin{table*}[!htbp] \footnotesize
\vspace{-3pt}
	\centering
	\caption{Quantitative results of different methods on four real-world
      benchmark datasets. The best results are indicated by \textcolor{red}{red}
      color and the second best results are denoted by \textcolor{blue}{blue} color. The results of `Average' are obtained by averaging the metric scores of all images from these four real-world datasets.}
      \vspace{-3pt}
	\setlength{\tabcolsep}{1mm}{
	\begin{tabular}{|C{.085\linewidth}|C{.085\linewidth}|C{.085\linewidth}|C{.085\linewidth}|C{.085\linewidth}|C{.085\linewidth}|C{.085\linewidth}|C{.085\linewidth}|C{.085\linewidth}|C{.085\linewidth}|}
		\hline
		\multirow{3}{*}{Dataset} & \multirow{3}{*}{Index} & \multicolumn{8}{c|}{Methods} \\ \cline{3-10}
		& & Input & LB14 & CEILNet & CEILNet & Zhang & BDN & BDN & ERRNet \\
		& & & \cite{Li2014Single} & \cite{fan2017generic} & F &\etal\cite{zhang2018single} & \cite{eccv18refrmv}& F & \\ \hline
\multirow{4}{*}{Real20} &PSNR&$19.05$&$18.29$&$18.45$&$20.32$&\textcolor{blue}{$21.89$}&$18.41$&$20.06$&\textcolor{red}{$22.89$}\\\cline{2-10}
&SSIM&$0.733$&$0.683$&$0.690$&$0.739$&\textcolor{blue}{$0.787$}&$0.726$&$0.738$&\textcolor{red}{$0.803$}\\\cline{2-10}
&NCC&$0.812$&$0.789$&$0.813$&$0.834$&\textcolor{red}{$0.903$}&$0.792$&$0.825$&\textcolor{blue}{$0.877$}\\\cline{2-10}
&LMSE&$0.027$&$0.033$&$0.031$&$0.028$&\textcolor{blue}{$0.022$}&$0.032$&$0.027$&\textcolor{red}{$0.022$}\\\hline
\multirow{4}{*}{Objects} &PSNR&$23.74$&$19.39$&$23.62$&$23.36$&$22.72$&$22.73$&\textcolor{blue}{$24.00$}&\textcolor{red}{$24.87$}\\\cline{2-10}
&SSIM&$0.878$&$0.786$&$0.867$&$0.873$&$0.879$&$0.856$&\textcolor{blue}{$0.893$}&\textcolor{red}{$0.896$}\\\cline{2-10}
&NCC&\textcolor{blue}{$0.981$}&$0.971$&$0.972$&$0.974$&$0.964$&$0.978$&$0.978$&\textcolor{red}{$0.982$}\\\cline{2-10}
&LMSE&$0.004$&$0.007$&$0.005$&$0.005$&$0.005$&$0.005$&\textcolor{blue}{$0.004$}&\textcolor{red}{$0.003$}\\\hline
\multirow{4}{*}{Postcard} &PSNR&$21.30$&$14.88$&$21.24$&$19.17$&$16.85$&$20.71$&\textcolor{red}{$22.19$}&\textcolor{blue}{$22.04$}\\\cline{2-10}
&SSIM&\textcolor{blue}{$0.878$}&$0.795$&$0.834$&$0.793$&$0.799$&$0.859$&\textcolor{red}{$0.881$}&$0.876$\\\cline{2-10}
&NCC&\textcolor{red}{$0.947$}&$0.929$&$0.945$&$0.926$&$0.886$&$0.943$&$0.941$&\textcolor{blue}{$0.946$}\\\cline{2-10}
&LMSE&$0.005$&$0.008$&$0.008$&$0.013$&$0.007$&$0.005$&\textcolor{red}{$0.004$}&\textcolor{blue}{$0.004$}\\\hline
\multirow{4}{*}{Wild} &PSNR&\textcolor{red}{$26.24$}&$19.05$&$22.36$&$22.05$&$21.56$&$22.36$&$22.74$&\textcolor{blue}{$24.25$}\\\cline{2-10}
&SSIM&\textcolor{red}{$0.897$}&$0.755$&$0.821$&$0.844$&$0.836$&$0.830$&\textcolor{blue}{$0.872$}&$0.853$\\\cline{2-10}
&NCC&\textcolor{red}{$0.941$}&$0.894$&$0.918$&$0.924$&$0.919$&\textcolor{blue}{$0.932$}&$0.922$&$0.917$\\\cline{2-10}
&LMSE&\textcolor{red}{$0.005$}&$0.027$&$0.013$&$0.009$&$0.010$&$0.009$&\textcolor{blue}{$0.008$}&$0.011$\\\hline
\multirow{4}{*}{\emph{Average}} &PSNR&$22.85$&$17.51$&$22.30$&$21.41$&$20.22$&$21.70$&\textcolor{blue}{$22.96$}&\textcolor{red}{$23.59$}\\\cline{2-10}
&SSIM&$0.874$&$0.781$&$0.841$&$0.832$&$0.838$&$0.848$&\textcolor{red}{$0.879$}&\textcolor{red}{$0.879$}\\\cline{2-10}
&NCC&\textcolor{blue}{$0.955$}&$0.937$&$0.948$&$0.943$&$0.925$&$0.951$&$0.950$&\textcolor{red}{$0.956$}\\\cline{2-10}
&LMSE&$0.006$&$0.011$&$0.009$&$0.010$&$0.007$&$0.007$&\textcolor{blue}{$0.006$}&\textcolor{red}{$0.005$}\\\hline

	\end{tabular}}
	\label{tb:method-com}
	\vspace{2pt}
\end{table*} 

\begin{figure*}
	\centering
\includegraphics[width=1\linewidth]{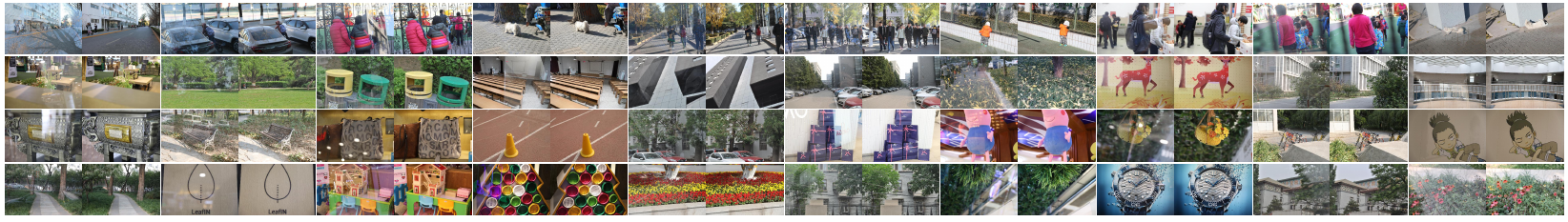}
\caption{Image samples in our unaligned image dataset. Our dataset covers a large variety of indoor and outdoor environments including dynamic scenes with vehicles, human, \etc.}
\label{fig:overview}
\end{figure*}

\subsection{Method Comparison on Benchmarks} \label{sec:model-com} 


In this section, we compare our ERRNet against state-of-the-art methods including the
optimization-based method of \cite{Li2014Single} (LB14) and the learning-based approaches
(CEILNet \cite{fan2017generic}, Zhang \etal \cite{zhang2018single}, and 
BDN \cite{eccv18refrmv}). For fair comparison, we
finetune these models on our training dataset and report results of both
the original pretrained model and finetuned version (denoted with a suffix '-F').

The comparison is conducted on four real-world datasets,
\ie 20 testing images in \cite{zhang2018single} and three sub-datasets from
SIR$^2$ \cite{Wan_2017_ICCV}. 
These three sub-datasets are 
captured under different conditions: (1) 20 controlled indoor scenes composed by
solid objects; (2) 20 different controlled scenes on postcards; and (3)
55 wild
scenes\footnote{Images indexed by 1, 2, 74 are removed due to misalignment.} with ground truth provided. In the following, we denote these datasets by `Real20', `Objects', `Postcard', and `Wild', respectively.




Table \ref{tb:method-com} summarizes the results of all competing
methods on four real-world datasets. The quality metrics include PSNR, SSIM \cite{wang2004image}, NCC
\cite{Xue2015ObstructionFree,Wan_2017_ICCV} and LMSE \cite{grosse2009ground}.
Larger values of PSNR, SSIM, and NCC indicate better performance, while a smaller value of LMSE implies a better result. Our ERRNet achieves the state-of-the-art
performance in `Real20' and `Objects' datasets. Meanwhile, our result is
comparable to the best-performing BDN-F on `Postcard' data. The quantitative
results on `Wild' dataset reveal a frustrating fact, namely, that no method
could outperform the naive baseline 'Input', suggesting that there is still large room
for improvement.

Figure \ref{fig:method-comparision} displays visual results on real-world
images. It can be seen that all compared methods fail to handle some strong
reflections, but our network more accurately removes many undesirable artifacts,
\eg removal of tree branches reflected on the building window in the fourth photo of Fig~\ref{fig:method-comparision}.

\begin{table*}[!htbp]
\centering
    \begin{tabular}{ l r r c c}
      \toprule
      Score Range Ratio & BDN-F & ERRNet &
      \multirow{5}{*}{\includegraphics[width=0.3\linewidth,clip,keepaspectratio]{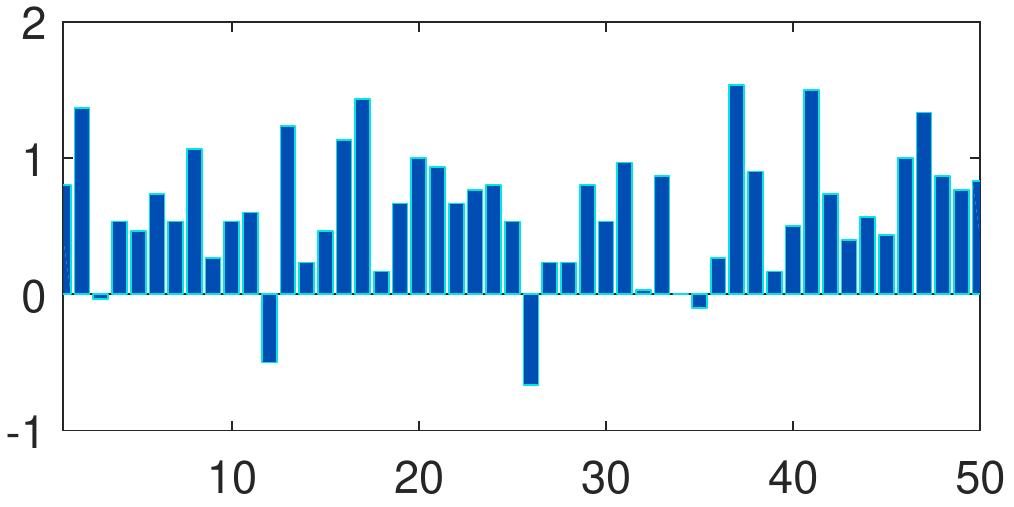}} &  
     \multirow{5}{*}{\includegraphics[width=0.3\linewidth,clip,keepaspectratio]{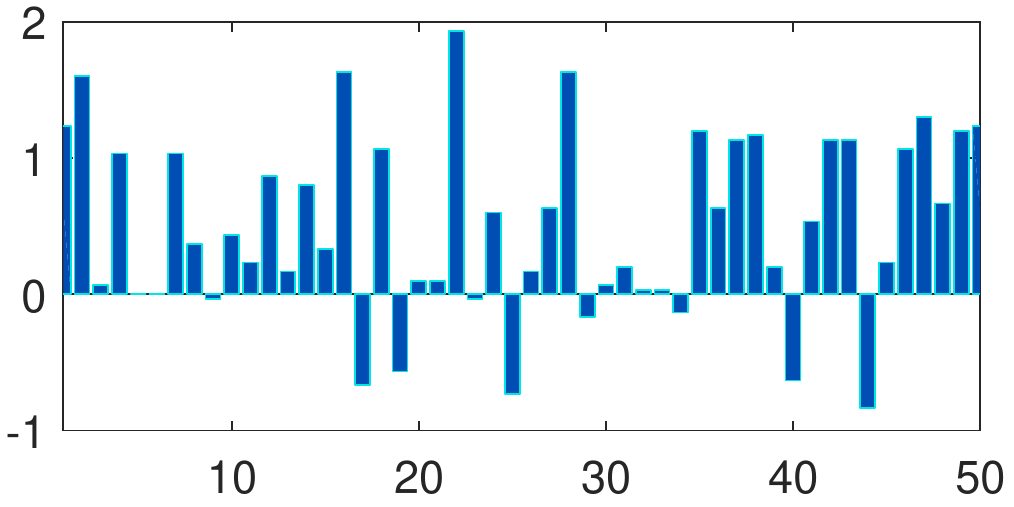}} \\
       $(0.25, 2]$  &  78\%  &  54\%  \\ \cline{1-3}
     $[-0.25, 0.25]$ & 18\% & 36\% \\ \cline{1-3}
    $[-2, -0.25)$ & 4\% & 10\% \\ \cline{1-3}
     
     \\
      Average Score &  0.62 & 0.51 \\ 
      \bottomrule
    \end{tabular}
    \caption{Human preference scores of self-comparsion experiments. Left:
      results of BDN-F; Right: results of ERRNet. X axis of each sub-figure  represents the image \# of testing images (50 in total).} \label{fig:MOS}
\end{table*}

\begin{figure*}[!htbp]
	\centering
	\setlength\tabcolsep{1pt}
	\begin{tabular}{cc|cc|cc}
  & & 	\multicolumn{2}{c|}{BDN-F}  & \multicolumn{2}{c}{ERRNet} \\
        	 \includegraphics[width=0.154\linewidth]{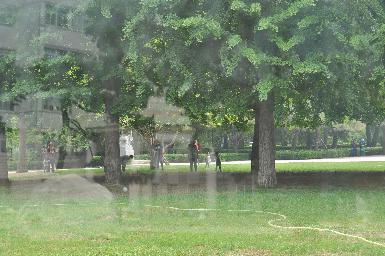}
		&  \includegraphics[width=0.154\linewidth]{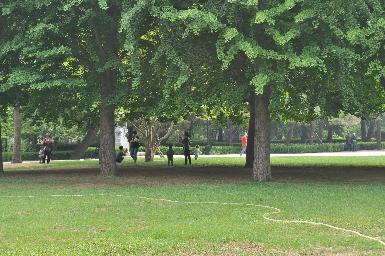}        	 
		&  \includegraphics[width=0.154\linewidth]{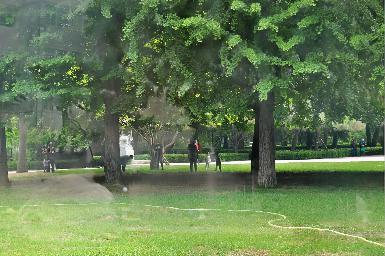}
		&  \includegraphics[width=0.154\linewidth]{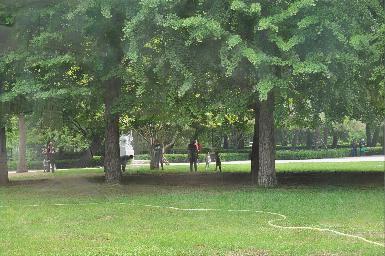}			 		
		&  \includegraphics[width=0.154\linewidth]{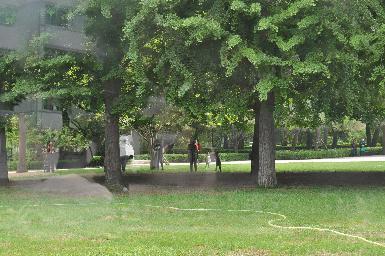}
		&  \includegraphics[width=0.154\linewidth]{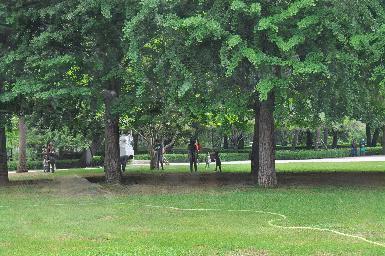} \\
		       \includegraphics[width=0.154\linewidth]{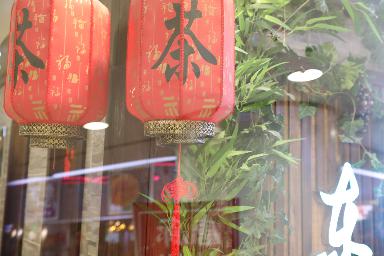}
		&  \includegraphics[width=0.154\linewidth]{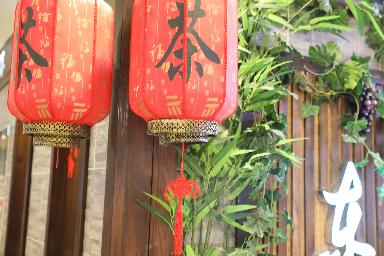}        	 		       
		&  \includegraphics[width=0.154\linewidth]{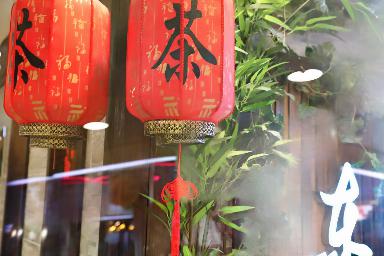}
		&  \includegraphics[width=0.154\linewidth]{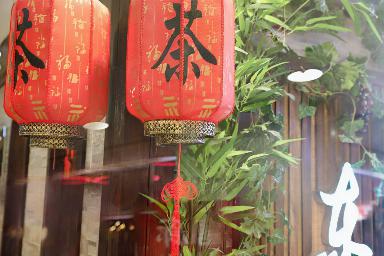}			       
		&  \includegraphics[width=0.154\linewidth]{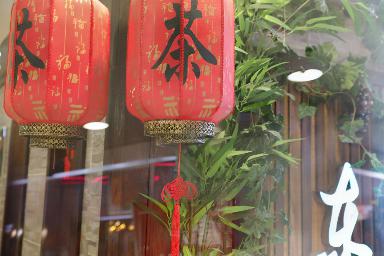}
		&  \includegraphics[width=0.154\linewidth]{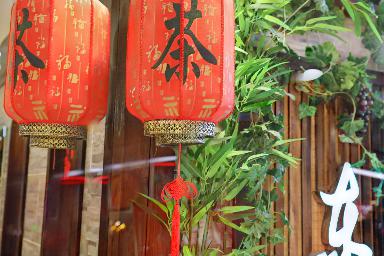}
		 \\				
		       \includegraphics[width=0.154\linewidth]{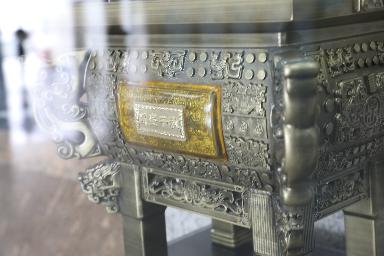}
		&  \includegraphics[width=0.154\linewidth]{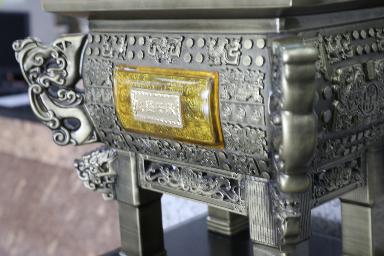}        	 		       
		&  \includegraphics[width=0.154\linewidth]{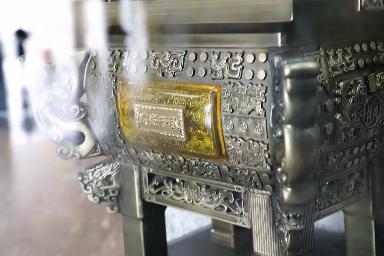}
		&  \includegraphics[width=0.154\linewidth]{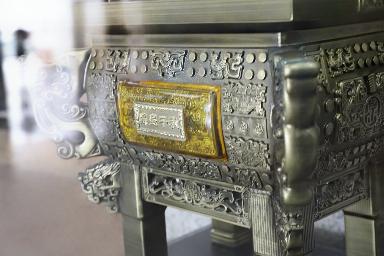}			 				       
		&  \includegraphics[width=0.154\linewidth]{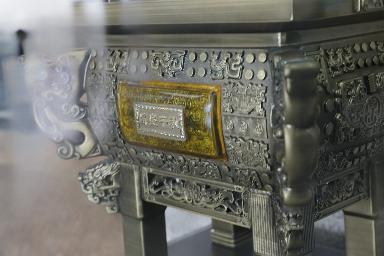}
		&  \includegraphics[width=0.154\linewidth]{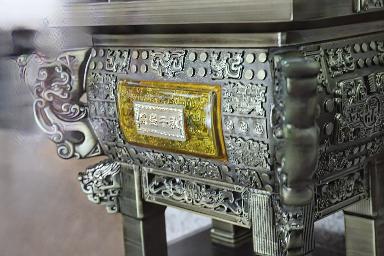} \\
		input & reference & w/o unaligned & w. unaligned   & w/o unaligned & w. unaligned  \\		
	\end{tabular} 
	\vspace{-5pt}
	\caption{Results of training with and without unaligned data. See \emph{suppl. material} for more examples. (\textbf{Best view on screen with zoom}) }
\label{fig:unaligned}
\vspace{-6pt}
\end{figure*}

\subsection{Training with Unaligned Data} \label{sec:unaligend-data}

To test our alignment-invariant loss on real-world unaligned
data, we first collected a dataset of unaligned image pairs with cameras and a portable glass, as shown in Fig.~\ref{fig:datacollection} . Both a DSLR camera and a smart phone are used to capture the images. We collected 450 image pairs in total, and some samples are shown in Fig~\ref{fig:overview}. These image pairs are randomly split into a training set of 400 samples and a testing set with 50 samples.

We conduct experiments on the BDN-F and ERRNet models, each of which is first trained on aligned dataset (w/o unaligned) as in Section~\ref{sec:model-com}, and then finetuned with our alignment-invariant loss and unaligned training data. The resulting pairs before and after finetuning are assembled for human assessment, as no existing numerical metric is available for evaluating unaligned data.

We asked 30 human observers to provide a preference score among \{-2,-1,0,1,2\} with 2 indicating the finetuned result is significantly better while -2 the opposite. To avoid bias, we randomly switch the image positions of each pair. In total, $3000$ human judgments are collected (2 methods, 30 users, 50 images pairs). More details regarding this evaluation process can be found in the \emph{suppl. material}.

Table \ref{fig:MOS} shows the average of human preference scores for the  resulting pairs of each method. As can be seen, human observers clearly tend to prefer the results produced by the finetuned models over the raw ones, which demonstrates the benefit of leveraging unaligned data for training independent of the network architecture.
Figure~\ref{fig:unaligned} shows some typical results of the two methods; the results are significantly improved by training on unaligned data.

\section{Conclusion}
We have proposed an enhanced reflection removal network together with an alignment-invariant loss function to help resolve the difficulty of single image reflection removal. We investigated the possibility to directly utilize misaligned training data, which can significantly alleviate the burden of capturing real-world training data. To efficiently extract the underlying knowledge from real training data, we introduce context encoding modules, which can be seamlessly embedded into our network to help discriminate and suppress the reflection component. Extensive experiments demonstrate our approach set a new state-of-the-art on real-world benchmarks of single image reflection removal, both quantitatively and visually.

{\small
\bibliographystyle{ieee}

}

\end{document}